\documentclass{article}

\usepackage{arxiv}

\usepackage[utf8]{inputenc} 
\usepackage[T1]{fontenc}    
\usepackage{hyperref}       
\usepackage{url}            
\usepackage{booktabs}       
\usepackage{amsfonts}       
\usepackage{nicefrac}       
\usepackage{microtype}      
\usepackage{cleveref}       
\usepackage{lipsum}         
\usepackage{graphicx}
\usepackage{natbib}
\usepackage{doi}
\usepackage[table]{xcolor}
\usepackage{graphicx}
\usepackage{subcaption}

\usepackage{authblk}


\hypersetup{
  colorlinks=true,
  linkcolor=cyan,
  citecolor=cyan,
  urlcolor=cyan
}

\title{Comparison of Different Deep Neural Network Models in the Cultural Heritage Domain}

\newif\ifuniqueAffiliation
\uniqueAffiliationfalse

\ifuniqueAffiliation 
\author{\hypersetup{urlcolor=red} \href{https://orcid.org/0000-0002-9554-2608}{\includegraphics[scale=0.06]{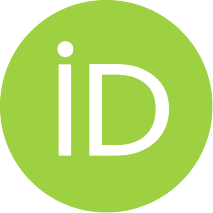}\hspace{1mm}Teodor Boyadzhiev} \\
	Institute of Mathematics and Informatics\\
	Bulgarian Academy of Sciences\\
	Sofia, Bulgaria \\
	\texttt{t.boyadzhiev@math.bas.bg} \\
	\And
	\href{https://orcid.org/0000-0003-4739-5778}{\includegraphics[scale=0.06]{orcid.pdf}\hspace{1mm}Gabriele Lagani} \\
	Institute of Information Science and Technologies \\
	Consiglio Nazionale delle Ricerche\\
	Pisa, Italy \\
	\texttt{gabriele.lagani@isti.cnr.it} \\
	\And
	\href{https://orcid.org/0000-0002-6985-0439}{\includegraphics[scale=0.06]{orcid.pdf}\hspace{1mm}Luca Ciampi} \\
	Institute of Information Science and Technologies \\
	Consiglio Nazionale delle Ricerche\\
	Pisa, Italy \\
	\texttt{luca.ciampi@isti.cnr.it} \\
	\And
	\href{https://orcid.org/0000-0003-0171-4315}{\includegraphics[scale=0.06]{orcid.pdf}\hspace{1mm}Giuseppe Amato} \\
	Institute of Information Science and Technologies \\
	Consiglio Nazionale delle Ricerche\\
	Pisa, Italy \\
	\texttt{giuseppe.amato@isti.cnr.it} \\
	\And
	\href{https://orcid.org/0000-0001-5056-7513}{\includegraphics[scale=0.06]{orcid.pdf}\hspace{1mm}Krassimira Ivanova} \\
	Institute of Mathematics and Informatics \\
	Bulgarian Academy of Sciences\\
	Sofia, Bulgaria \\
	\texttt{kivanova@math.bas.bg} \\
}
\else
\usepackage{authblk}

\newbox{\orcid}\sbox{\orcid}{\includegraphics[scale=0.06]{orcid.pdf}} 

\author[1]{\href{https://orcid.org/0000-0002-9554-2608}{\usebox{\orcid}\hspace{1mm}Teodor Boyadzhiev\thanks{\texttt{t.boyadzhiev@math.bas.bg}}}}
\author[2]{\href{https://orcid.org/0000-0003-4739-5778}{\usebox{\orcid}\hspace{1mm}Gabriele Lagani\thanks{\texttt{gabriele.lagani@isti.cnr.it}}}}
\author[2]{\href{https://orcid.org/0000-0002-6985-0439}{\usebox{\orcid}\hspace{1mm}Luca Ciampi\thanks{\texttt{luca.ciampi@isti.cnr.it}}}}
\author[2]{\href{https://orcid.org/0000-0003-0171-4315}{\usebox{\orcid}\hspace{1mm}Giuseppe Amato\thanks{\texttt{giuseppe.amato@isti.cnr.it}}}}
\author[1]{\newline\href{https://orcid.org/0000-0001-5056-7513}{\usebox{\orcid}\hspace{1mm}Krassimira Ivanova\thanks{\texttt{kivanova@math.bas.bg}}}}

\affil[1]{Institute of Mathematics and Informatics, Bulgarian Academy of Sciences, Sofia, Bulgaria}
\affil[2]{Institute of Information Science and Technologies, Consiglio Nazionale delle Ricerche, Pisa, Italy}
\fi

\renewcommand{\shorttitle}{Comparison of Different Deep Neural Network Models in the Cultural Heritage Domain}

\hypersetup{
pdftitle={A template for the arxiv style},
pdfsubject={q-bio.NC, q-bio.QM},
pdfauthor={David S.~Hippocampus, Elias D.~Striatum},
pdfkeywords={First keyword, Second keyword, More},
}

\begin{document}
\maketitle

\begin{abstract}
The integration of computer vision and deep learning is an essential part of documenting and preserving cultural heritage, as well as improving visitor experiences. In recent years, two deep learning paradigms have been established in the field of computer vision: convolutional neural networks and transformer architectures. The present study aims to make a comparative analysis of some representatives of these two techniques of their ability to transfer knowledge from generic dataset, such as ImageNet, to cultural heritage specific tasks. The results of testing examples of the architectures VGG, ResNet, DenseNet, Visual Transformer, Swin Transformer, and PoolFormer, showed that DenseNet is the best in terms of efficiency-computability ratio.
\end{abstract}


\section{Introduction}
In the dynamic field of computer vision, the recognition of monuments through image analysis has evolved as a compelling application, particularly within the context of cultural heritage \cite{DBLP:journals/jocch/AmatoFG15}. With the advent of deep learning (DL), there has been a transformative leap in the ability to automatically discern and categorize iconic structures with unprecedented accuracy and efficiency. Indeed, these techniques, such as convolutional neural networks (CNNs) and Transformer, have demonstrated remarkable capabilities in extracting complex features from images, enabling the development of effective models for monument recognition.

The integration of computer vision and DL in monument recognition not only facilitates the automated identification of monuments but also contributes significantly to the documentation and safeguarding of cultural heritage \cite{DELIGIORGI2021102787}. By automating the process of cataloguing and classifying monuments, these technologies assist in the creation of comprehensive databases, aiding historians, archaeologists, and conservationists in their endeavours. On the other side, the advent of mobile applications equipped with monument recognition capabilities enhanced visitors' experience, empowering tourists to actively participate in the classification and documentation of monuments as they explore historical sites \cite{isprs-archives-XLII-2-W15-77-2019}.

In this work, we conduct a comparative experimental analysis of popular DL architectures based on CNNs and Transformers for monument recognition in the cultural heritage domain. Starting from pre-trained publicly available models trained upon extensive image datasets like ImageNet \cite{DBLP:conf/cvpr/DengDSLL009}, our method involves implementing a fine-tuning strategy tailored to the specific task of monument classification. Notably, the capability to adapt to new tasks using only a sparse set of available training samples is critical, as gathering significant amounts of annotated data proves prohibitively expensive. Specifically, we consider VGG-11 \cite{DBLP:journals/corr/SimonyanZ14a}, ResNet-34 \cite{DBLP:conf/cvpr/HeZRS16}, DenseNet-121 \cite{DBLP:conf/cvpr/HuangLMW17}, ViT-S \cite{DBLP:conf/iclr/DosovitskiyB0WZ21}, Swin-T \cite{DBLP:conf/iccv/LiuL00W0LG21}, and PoolFormer-S24 \cite{DBLP:conf/cvpr/YuLZSZWFY22} architectures, and we systematically compare their outcomes to monument classification in terms of accuracy and computational cost. We fine-tune and test these models over the Pisa Dataset \cite{DBLP:journals/jocch/AmatoFG15}, a small collection of 1227 images that includes 12 cultural heritage sites or monuments in Pisa, Italy. 

The rest of this paper is organized as follows. Section \ref{sec:related} contains a short review of tested architectures. Section \ref{sec:dataset} describes characteristics of the Pisa dataset. Section \ref{sec:experiments} is devoted to experimental settings. In Section \ref{sec:results}, the achieved results are analysed. Finally, some conclusions and directions for further research are pointed out in Section \ref{sec:conclusion}.

\section{Tested Architectures}
\label{sec:related}
Past years of research in the domain of computer vision established two essential paradigms for deep learning solutions: Convolutional Neural Network (CNN) and Transformer architectures. The first type of architecture is known to have a relatively low computational footprint, while also being able to generalize easily in image-related problems. This is due to the translation-invariance and locality properties of convolutional layers, which translate to an effective inductive bias, useful for generalization. More recently, Transformer models gained popularity, bringing the ability to capture relationships in the input on a global scale, thanks to their attention-based processing. Unfortunately, the computational complexity of attention layers grows quadratically with the input size, making these models poorly scalable to real-world applications. Some variants of Transformer models have been proposed to address this problem \cite{DBLP:journals/csur/TayDBM23}, replacing full attention blocks with more efficient alternatives, without reducing performance. Transformer-based architectures are known to be less prone to generalization and often require pre-training on very large datasets to achieve satisfactory results \cite{DBLP:conf/emnlp/CsordasIS21}. Fortunately, pre-trained models of both types of paradigms are publicly available online and can be fine-tuned on task-specific datasets and deployed in a desired applicative scenario.

Therefore, in this work, to address the task of monument recognition for cultural heritage applications, we consider some popular architectures belonging both to the category of CNNs, and to that of Transformers, and we conduct a comparative experimental analysis of such models in this scenario. We used a pre-trained model, built on ImageNet and we pursued a fine-tuning strategy on the specific monument classification task, comparing the effectiveness and transferability of such pre-trained features on the new task. Indeed, the capacity to adapt to the new task from few available training samples is crucial, because gathering consistent amounts of annotated data is costly. 

\begin{table}[t]
    \centering
    \normalsize
    \caption{Comparison of different models in terms of number of parameters and computing cost (in GFLOPs). Best values are highlighted in bold.}
    \label{tab:comparison}
    \renewcommand{\arraystretch}{1.2}
    \rowcolors{2}{gray!10}{white}
    \begin{tabular}{lcc}
        \toprule
        \textbf{Model} & \textbf{\# Parameters} & \textbf{GFLOPs} \\
        \midrule
        VGG-11           & 113M & 7.60 \\
        ResNet-34        & 64M  & 3.68 \\
        DenseNet-121     & \textbf{8M}   & \textbf{2.91} \\
        ViT-S            & 22M  & 4.25 \\
        Swin-T           & 28M  & 4.50 \\
        PoolFormer-S24   & 21M  & 3.41 \\
        \bottomrule
    \end{tabular}
\end{table}

Specifically, the models that we consider in our comparison are the following:
\begin{itemize}
    \item VGG-11 \cite{DBLP:journals/corr/SimonyanZ14a} is a purely convolutional architecture with 11 layers and 113M parameters;
    \item ResNet-34 \cite{DBLP:conf/cvpr/HeZRS16} is an evolution of purely convolutional architectures which introduces residually connected blocks, allowing to stabilize gradient backpropagation and model training even with deeper architectures; it has 34 layers and 64M parameters;
    \item Dense Network – DenseNet \cite{DBLP:conf/cvpr/HuangLMW17} further enhances the residual connection structure by adding skip connections from each layer to every subsequent layer, thus further empowering the representation power and gradient propagation abilities of DenseNet models; in particular, DenseNet-121 uses 121 layers and 8M parameters;
    \item Vision Transformer – ViT \cite{DBLP:conf/iclr/DosovitskiyB0WZ21} encodes images as a collection of tokens, extracted from non-overlapping image patches of fixed size, and then applies a series of Transformer encoder blocks on these tokens, thus leveraging a Multi-Head Self-Attention mechanism to model global relationships among tokens; ViT-S version has 12 layers and 22M parameters;
    \item Swin Transformer \cite{DBLP:conf/iccv/LiuL00W0LG21} is a state-of-the-art Transformer-based architecture which leverages attentional processing on local windows, whose location is shifted from one layer to the next, to allow information to propagate also along adjacent windows; the Tiny version (Swin-T) of the architecture contains 12 Transformer blocks (implemented leveraging the Swin attention mechanism described so far), and 28M parameters;
    \item PoolFormer \cite{DBLP:conf/cvpr/YuLZSZWFY22} is another state-of-the-art architecture based on the Transformer structure, which instead replaces attention blocks with a simple average pooling, experimentally showing that the performance of the trained models remains comparable to that of purely attention-based models; PoolFormer-S24 version amounts to 24 layers, and 21M parameters.
\end{itemize}

Table \ref{tab:comparison} summarizes the number of parameters and computational footprint (in terms of GFLOPs) of each model.

\begin{table}[t]
    \centering
    \normalsize
    \caption{Pisa dataset – categories and number of examples.}
    \label{tab:dataset}
    \renewcommand{\arraystretch}{1.2}
    \rowcolors{2}{gray!10}{white}
        \begin{tabular}{ll|ll}
        \toprule
        \textbf{Category} & \textbf{\# Photos} & \textbf{Category} & \textbf{\# Photos} \\
        \midrule
        Battistero (the baptistery of St. John) & 104 & Basilica of San Piero (church)     & 48  \\
        Duomo (the cathedral of St. Mary)       & 130 & Certosa (the charterhouse)         & 53  \\
        Leaning Tower (campanile)               & 119 & Chiesa della Spina (Gothic church) & 112 \\
        Camposanto Monumentale – exterior       & 46  & Guelph Tower                        & 71  \\
        Camposanto Monumentale – portico        & 138 & Palazzo della Carovana (building)  & 101 \\
        Camposanto Monumentale – field          & 113 & Palazzo dell’Orologio (building)   & 92  \\
        \bottomrule
        \end{tabular}
\end{table}

\begin{figure}[t]
    \centering
    \begin{subfigure}[b]{0.13\textwidth}
        \includegraphics[width=\linewidth]{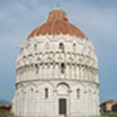}
    \end{subfigure}
    \begin{subfigure}[b]{0.13\textwidth}
        \includegraphics[width=\linewidth]{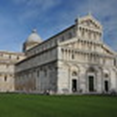}
    \end{subfigure}
    \begin{subfigure}[b]{0.13\textwidth}
        \includegraphics[width=\linewidth]{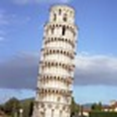}
    \end{subfigure}
    \begin{subfigure}[b]{0.13\textwidth}
        \includegraphics[width=\linewidth]{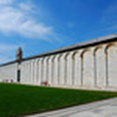}
    \end{subfigure}
    \begin{subfigure}[b]{0.13\textwidth}
        \includegraphics[width=\linewidth]{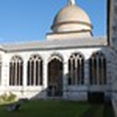}
    \end{subfigure}
    \begin{subfigure}[b]{0.13\textwidth}
        \includegraphics[width=\linewidth]{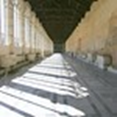}
    \end{subfigure}

    \vspace{0.5em}

    \begin{subfigure}[b]{0.13\textwidth}
        \includegraphics[width=\linewidth]{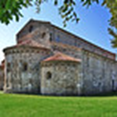}
    \end{subfigure}
    \begin{subfigure}[b]{0.13\textwidth}
        \includegraphics[width=\linewidth]{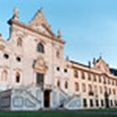}
    \end{subfigure}
    \begin{subfigure}[b]{0.13\textwidth}
        \includegraphics[width=\linewidth]{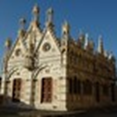}
    \end{subfigure}
    \begin{subfigure}[b]{0.13\textwidth}
        \includegraphics[width=\linewidth]{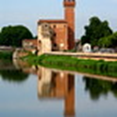}
    \end{subfigure}
    \begin{subfigure}[b]{0.13\textwidth}
        \includegraphics[width=\linewidth]{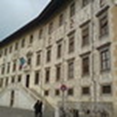}
    \end{subfigure}
    \begin{subfigure}[b]{0.13\textwidth}
        \includegraphics[width=\linewidth]{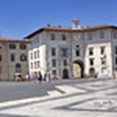}
    \end{subfigure}

    \caption{Examples from the Pisa dataset.}
    \label{fig:dataset_sample}
\end{figure}

\section{Dataset Description}
\label{sec:dataset}
In the experiments executed in this paper, we used the Pisa Dataset \cite{DBLP:journals/jocch/AmatoFG15}. The dataset contains 1227 photos of 12 cultural heritage sites or monuments in Pisa, Italy. The photos came from the online photo service Flickr. The IDs and labels of the photos we used for these experiments can be downloaded from \href{https://falchi.isti.cnr.it/pisaDataset/}{https://falchi.isti.cnr.it/pisaDataset/}. 

Table \ref{tab:dataset} shows some statistics of the dataset while Figure \ref{fig:dataset_sample} depicts some samples for each category.

\section{Experimental Setting}
\label{sec:experiments}
Each network architecture was pre-trained on ImageNet, and it was fine-tuned on the Pisa dataset. Each fine-tuning was repeated 20 times with random data splits to find confidence intervals on the results. 

To test whether the fine-tuning is successful, a baseline classification was used for comparison. The baseline classification consists of feature extraction using the pre-trained networks, without fine-tuning, and then classification using the \textit{k}-nearest neighbour classifier in the latent space.

All the experiments are performed using torch v2.0 and torchvision v1.16 and augmentation is done using torchvision v2 API. The number of floating-point operations is counted using fvcore v0.1.5. The \textit{k}-nearest neighbour classifier was provided by the library scikit-learn 1.4.1.

\paragraph{Data Preparation.} The initial dataset contained 1127 JPG images. Broken images and images which have shorter sides of less than 200 pixels were removed. Then the images from each category were split into 80\% for training and testing and 20\% for validation. When fine-tuning a network, the data allocated for training and testing was randomly split into 80\% for training and the remaining 20\% for testing. Each trail has 64\% of the total data for training, 16\% for testing and 20\% for validation, where the training and testing splits are different, and the validation split is the same.

\paragraph{Baseline.} The results of each fine-tuning trail are compared with a baseline. For each trail, the images in the training, testing, and validation split were used to extract latent features. The features are extracted by using the corresponding pre-trained network on ImageNet and removing the last fully connected layer. The images are transformed in the same way as the testing augmentation, described below. The training features were used for building \textit{k}-nearest neighbour classifier and the accuracy of the classifier was measured on the testing and validation features. The number of neighbours was set to 5, based on trials-and-error.

\paragraph{Fine Tuning.} Each network was initialised with weights pre-trained on ImageNet except for the last fully connected layer, which was replaced with randomly initialised weights.

\paragraph{Data augmentation.} Each image was represented in RGB colour space and scaled between 0 and 1. For training augmentation, each image was randomly scaled so that the shorter edge to be between 256 and 480 pixels while preserving the aspect ratio. This operation is followed by a random crop of a window of 224x224 pixels and then AutoAugment \cite{DBLP:conf/cvpr/CubukZMVL19} with ImageNet policy. For testing augmentation, a fixed shorter edge resize to 256 pixels was used, followed by a centre crop of 224x224 pixels. In training and testing, each image was normalised using mean [0.485, 0.456, 0.406] and standard deviation of [0.229, 0.224, 0.225], which was consistent with ImageNet normalisation.

\paragraph{Training Algorithm.} Each network was finetuned using the AdamW optimiser with a learning rate of $10^{-4}$ with cosine annealing and weight decay of $10^{-2}$ for 300 epochs. The batch size was set to 64 and the loss is cross-entropy with label smoothing set to 0.1.

\section{Results}
\label{sec:results}
\begin{figure}[t]
    \centering
    \includegraphics[width=0.9\textwidth]{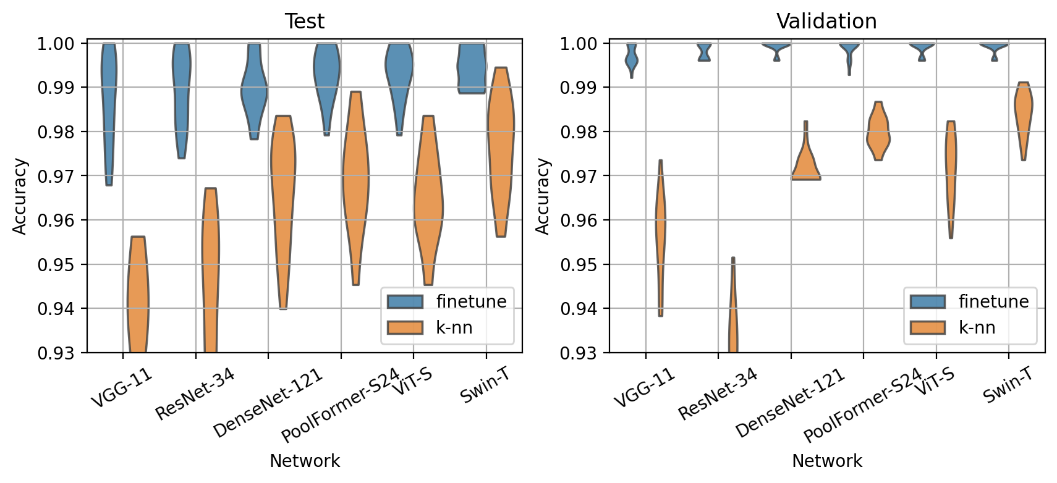}
    \caption{Accuracy measured on the testing and validation data for each fine-tuned network, shown in blue, compared to the \textit{k}-nearest neighbour baseline classification, shown in orange, for all 20 trails.}
    \label{fig:results1}
\end{figure}

Figure \ref{fig:results1} shows the distribution of training and testing accuracy across all the trails for the fine-tuned networks and the \textit{k}-nearest neighbours. The results show a significant improvement in accuracy when fine-tuning is used over \textit{k}-nearest neighbour classification. 

\begin{figure}[htbp]
    \centering
    \includegraphics[width=0.9\textwidth]{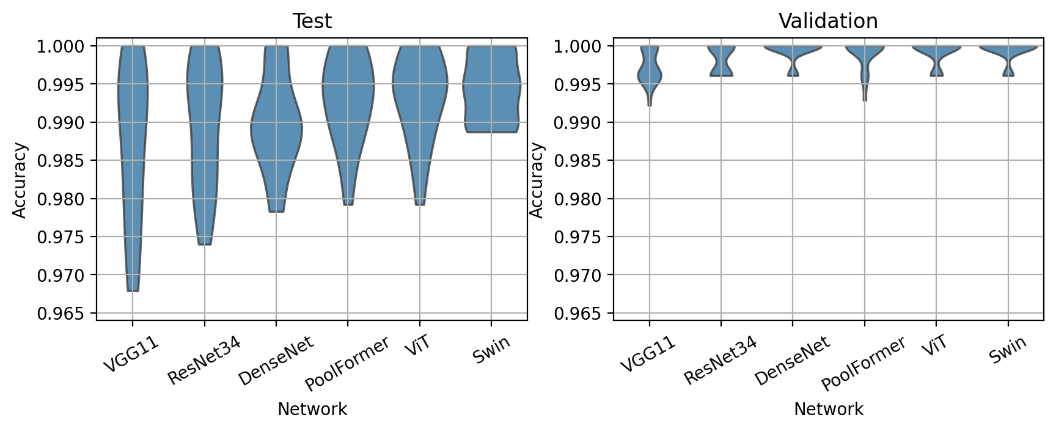}
    \caption{Accuracy measured on the testing and validation data for each fine-tuned network for all 20 trails.}
    \label{fig:results2}
\end{figure}

Figure \ref{fig:results2} shows the distributions of the testing and validation accuracy across all the trails only for the fine-tuned networks. The testing results show that VGG-11 and ResNet-34 performed worse than DenseNet-121, PoolFormer-S24 and ViT-S. The best testing accuracy was recorded by the Swin-T. The validation results showed that the worst-performing networks were VGG-11 and PoolFormer-S24. ResNet-34 had better-case accuracy than the other convolutional networks, however, the distribution has more density towards the lower accuracy. DenseNet-121, ViT-S and Swin-T have very similar validation accuracy and can be regarded as equal quality. However, from these 3 networks, the DenseNet-121 uses the least computations to process an image at around 2.9GFlop, while ViT-S makes approximately 4.25GFlop and Swin-T makes 4.5GFlop.

Overall Swin-T has the best testing accuracy, however, the validation results show that DenseNet is preferable, since it is significantly more efficient and has the same validation accuracy as Swin-T.

\section{Conclusion}
\label{sec:conclusion}
The comparative analysis of popular DL architectures over the Pisa Dataset shows the ability of each architecture to adapt to a new domain with limited training data in terms of accuracy and computational cost. The experimental results reveal significant improvements in accuracy achieved through fine-tuning compared to baseline \textit{k}-nearest neighbour classification. Notably, Swin-T emerges as the top-performing model in terms of testing accuracy, while DenseNet-121 exhibits favourable computational efficiency while having the same validation accuracy as Swin-T. This suggests a nuanced trade-off between accuracy and computational cost, with DenseNet-121 being a preferable choice for efficiency considerations. Overall, this study confirms the efficacy of applying such DL architectures in monument recognition and lays the groundwork for further research in this domain. 
We intend to continue the research by studying the possibilities of combining the supervised fine-tuning with Hebbian pre-training. The advantages of the Hebbian learning rule are that it is unsupervised which allows for easier data collection. The Hebbian learning algorithm usually converges within one iteration over the dataset. Also biologically inspired techniques, such as Hebbian learning, typically have better generalisation abilities \cite{DBLP:conf/cogsci/OrorbiaM23a}. We expect that combining the advantages of Hebbian learning with the qualities of CNNs and Transformers will lead to efficient models for monument recognition.

\section{Acknowledgments}
This research is partially financed by a bilateral program between the Bulgarian Academy of Sciences and Consiglio Nazionale delle Ricerche, Italy, under the project “Innovative Semi-Supervised Learning Techniques for Enhancing Cultural Heritage Discovery”.

\bibliographystyle{unsrtnat}
\bibliography{references}  






\end{document}